# DRAformer: Differentially Reconstructed Attention Transformer for Time-Series Forecasting

Benhan Li, Shengdong Du*, Tianrui Li, Jie Hu, Zhen Jia
Southwest Jiaotong University, Chengdu, China
bhli@my.swjtu.edu.cn, {sddu, trli, jiehu, zjia}@swjtu.edu.cn

## ABSTRACT

Time-series forecasting plays an important role in many real-world scenarios, such as equipment life cycle forecasting, weather forecasting, and traffic flow forecasting. It can be observed from recent research that a variety of transformer-based models have shown remarkable results in time-series forecasting. However, there are still some issues that limit the ability of transformer-based models on time-series forecasting tasks: (i) learning directly on raw data is susceptible to noise due to its complex and unstable feature representation; (ii) the self-attention mechanisms pay insufficient attention to changing features and temporal dependencies. In order to solve these two problems, we propose a transformer-based differentially reconstructed attention model DRAformer. Specifically, DRAformer has the following innovations: (i) learning against differenced sequences, which preserves clear and stable sequence features by differencing and highlights the changing properties of sequences; (ii) the reconstructed attention: *integrated distance attention* exhibits sequential distance through a learnable Gaussian kernel, *distributed difference attention* calculates distribution difference by mapping the difference sequence to the adaptive feature space, and the combination of the two effectively focuses on the sequences with prominent associations; (iii) the reconstructed decoder input, which extracts sequence features by integrating variation information and temporal correlations, thereby obtaining a more comprehensive sequence representation. Extensive experiments on four large-scale datasets demonstrate that DRAformer outperforms state-of-the-art baselines.

## 1 INTRODUCTION

Time-series forecasting is applied in many important fields, such as sensor parameter forecasting, weather forecasting, traffic flow forecasting. Over the past few decades, researchers have proposed a variety of time-series forecasting methods. One of them is based on statistical learning, such as autoregressive moving average model (ARIMA) [1], support vector machine (SVM) [2], artificial neural networks (ANN) [3], etc. With the advent of the era of big data, traditional shallow learning models cannot effectively learn nonlinear features of large-scale sequences. The other is based on deep learning. The family of recurrent neural networks ([4], [5]) is an effective method for processing time-series, which characterizes historical information in predictions, however, it pays insufficient attention to global information and is prone to gradient vanishing. For such problems, the parallel training model transformer was proposed [6] and quickly demonstrated amazing capabilities in several important domains. Recently, researchers have been keen to develop transformer-based models for time-series forecasting tasks, and significant progress has been made. Specifically, existing work aims to reduce computational complexity and improve multi-step prediction capabilities for long sequences. For example, Reformer [7] uses locality-sensitive hashing dot-product attention, which changes the complexity from $O(L^2)$ to $O(L\log L)$. LogTrans [8] uses *LogSparse* Transformer to reduce memory overhead to $O(L(\log L)^2)$. Informer [9] proposes the *ProbSparse* self-attention mechanism, which achieves $O(L\log L)$ in terms of complexity and memory usage. Autoformer [10] designs a new decomposition architecture with an *Auto-Correlation* mechanism, which enables the model to have the ability of progressive decomposition of complex sequences. FEDformer [11] combines the transformer with the seasonal-trend decomposition method, enabling the model to capture the global profile of time series. However, there are still some issues that limit the ability of transformer-based models for time-series forecasting:

(i) The forecasting process is susceptible to data fluctuations. Due to the accumulation of errors at the perturbed points, the predictive ability of the model will drop sharply with the increase of the number of prediction steps, which reduces the robustness of the model.

(ii) Insufficient attention to the changing characteristics and time dependence of the sequence. The dot-product form of transformer-based models focuses on capturing global importance information, which does not pay enough attention to relative changes in sequences. Furthermore, the high-dimensional parameter matrix in the attention mechanism dilutes the temporal correlation and increases the computational complexity.

In order to solve these two problems, we propose a differential reconstruction attention transformer, called DRAformer. Specifically, it contains the following innovations:

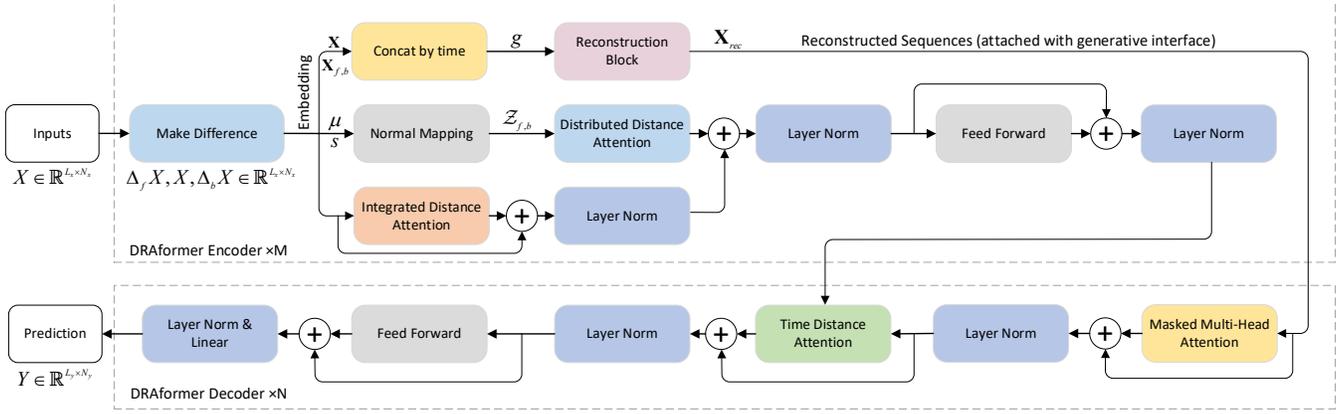

Figure 1: The overall architecture of DRAformer. The encoder builds relative attention by learning the integrated distance and distribution difference of the differential sequence, and the reconstruction block extracts the long-term dependencies of the sequence. The generative interface is used to generate all predictions at once, and the decoder further fuses temporal information (green modules in the figure) based on the changing features extracted by the encoder, with fewer training parameters.

(i) The first-order difference of the original sequence is used as the common input of the encoder, which highlights the stable hidden features of the sequence and introduces the variability into the learning process, which makes the model have better interpretability.

(ii) The attention form is reconstructed for the difference sequence. Specifically, integrated distance attention calculates relative attention scores based on temporal order and sequence changes, and distributed difference attention calculates relative change probabilities by mapping sequences into distribution with approximately zero mean. The combination of the two enables the model to deeply learn sequence correlations, and reduce the number of parameters that need to be trained.

(iii) The reconstructed decoder input. We utilize convolutional networks and max-pooling to extract temporal features, and set a sliding weight window to extract embedded features and record historical information. This allows the model to preserve long-term serial dependencies. Furthermore, we improve long-term prediction speed and reduce error accumulation with one-shot generative inference.

Experiments on multiple large-scale datasets show that DRAformer has excellent long-term prediction ability.

## 2 PRELIMINARY

For the input sequence $X = \{x_1, x_2, ..., x_{L_x} | x_i \in \mathbb{R}^{N_x}\}$, our goal is to predict several future values $Y = \{y_1, y_2, ..., y_{L_y} | y_i \in \mathbb{R}^{N_y}\}$ associated with it, where $N_x, N_y$ is the number of variables, and $L_x, L_y$ is the sequence length. Furthermore, we denote the forward first difference as $\Delta_f X^t = X^{t+1} - X^t$ and the backward first difference as $\Delta_b X^t = X^t - X^{t-1}$. Embedding is a linear map function, $d_{model}$ is the dimension of the embedding, and the result of embedding is $\mathbf{X}_f = \Delta_f X W_f$, $\mathbf{X} = X W_x$, $\mathbf{X}_b = \Delta_b X W_b$, where $W_f, W_x, W_b \in \mathbb{R}^{N_x \times d_{model}}$. In this work, we mainly design relative attention for the differential form because it has a more stable feature representation.

## 3 METHODOLOGY

The architecture of DRAformer is shown in Figure 1. In this section, we will focus on the specific methods of reconstructing attention and how the reconstructed sequence can effectively capture long-term dependencies.

### 3.1 Differential Encoder with Reconstructed Attention

*3.1.1 Integrated distance attention* Considering the temporal distance and sequence distance, we apply a learnable Gaussian kernel to compute the synthetic distance attention score:

$$IDA = \text{Softmax}\left[\frac{1}{\sqrt{2\pi}\sigma_i} \exp\left(\frac{-|i-j|^2 dist_{i,j}^2}{2\sigma_i^2}\right)_{i,j\in\{1,2,...,L_x\}}\right]\mathcal{V}_x \quad (1)$$

where $IDA \in \mathbb{R}^{L_x \times d_{model}}$, $\mathcal{V}_x = \mathbf{X}W_v \in \mathbb{R}^{L_x \times d_{model}}$, $|i-j|$ is the temporal distance between the i-th and the j-th time point, $dist$ is the sequence distance, $\sigma = XW_\sigma \in \mathbb{R}^{L_x}$, and Softmax(·) is the activation function that normalizes the results by row to get an approximate probability distribution. Since Mahalanobis distance (MD) is not affected by the dimension, we use MD with regularization to represent $t$:

$$MD_{i,j}^2 = (\Delta_f X^i - \Delta_b X^j)(\Sigma + \lambda I)^{-1}(\Delta_f X^i - \Delta_b X^j)^T \quad (2)$$

where $\Sigma$ is the covariance matrix under the prior condition that $\Delta_f X$ and $\Delta_b X$ are identically distributed, $I$ is the identity matrix of the order $L_x$, and $\lambda$ is the scaling factor.

*3.1.2 Distributed difference attention* To measure the difference in probability distributions between different series, we first map the difference series to an approximately normal distribution by a learnable function as follows:

$$\mathcal{Z}(X; \Sigma) = \frac{X - \mu(X;\Sigma)}{s(X;\mu,\Sigma) + \varepsilon} \quad (3)$$

where $\mu(X;\Sigma) = X\varphi(\Sigma W_\mu)$, $s(X;\mu,\Sigma) = \sqrt{(X-\mu)^2 \varphi(\Sigma W_s)}$, $Z(X;\Sigma) \in \mathbb{R}^{L_x \times N_x}$. $\Sigma$ is the covariance matrix from 3.1.1, $\varepsilon$ is an infinitesimal quantity, $W_\mu, W_s \in \mathbb{R}^{N_x}$, and $\varphi = \text{Softmax}(\cdot)$ is the activation function that rescales the weights to [0,1] to fit the probability distribution. The Jensen-Shannon divergence makes up for the asymmetry of the Kullback-Leibler divergence. We denote the Jensen-Shannon divergence matrix, which measures the difference in distributions, as:

$$\mathcal{J}_{i,j} = \frac{1}{2}\text{KL}\left[\varphi(\mathcal{Z}_f^i) || \left(\frac{\mathcal{Z}_f^i + \mathcal{Z}_b^j}{2}\right)\right] + \frac{1}{2}\text{KL}\left[\varphi(\mathcal{Z}_b^j) || \left(\frac{\mathcal{Z}_f^i + \mathcal{Z}_b^j}{2}\right)\right] \quad (4)$$

where i, j $\in \{1,2,\ldots,L_x\}$, $\mathcal{Z}_f^i = \mathcal{Z}(\Delta_f X^i; \Sigma)$, $\mathcal{Z}_b^i = \mathcal{Z}(\Delta_b X^j; \Sigma)$, $\text{KL}[A||B] = \sum_n A(n)\log[A(n)/B(n)]$ is the Kullback-Leibler divergence between the two sequences, where n is the variable position. $\varphi = \text{Softmax}(\cdot)$ maps the sequence to [0,1]. The range of Jensen-Shannon divergence values is [0,1], which is similar to the function $\text{Softmax}(QK^T/\sqrt{d})$ in transformer and corresponds to a probability distribution. From this, we express the distributional difference attention score as:

$$JSA = \mathcal{J}\mathcal{V}_f + \mathcal{J}^T \mathcal{V}_b \quad (5)$$

where $\mathcal{V}_f = \mathcal{Z}_f W_f$, $\mathcal{V}_b = \mathcal{Z}_b W_b$, and $W_f, W_b \in \mathbb{R}^{d_{model} \times d_{model}}$.

## 3.2 Reconstructed Decoder Input

In order to make the reconstructed decoder input sequence contain sufficient variation features, we first concatenate the encoded original sequence and the difference sequence in the time dimension:

$$g_t = \text{Concat}(\mathbf{X}_f^t, \mathbf{X}^t, \mathbf{X}_b^t) \in \mathbb{R}^{3 \times d_{model}} \quad (6)$$

Then, as shown in Figure 1, the following two ways are used to extract features:

*3.2.1 Time distillation* We use 1-D convolution and max-pooling to perform a "distillation" operation in the temporal dimension to extract temporal information, and the reconstructed sequence is as follows:

$$\mathcal{T} = \text{Maxpool1d}(\text{ELU}(\text{Conv1d}[g])) \quad (7)$$

where $\in \mathbb{R}^{L_x \times k}$, $k$ is the number of features to extract. Conv1d($\cdot$) represents one-dimensional convolution in the time dimension (kernel size=3, stride=3). ELU($\cdot$) is the activation function and Maxpool1d($\cdot$) is max-pooling in the embedding dimension.

*3.2.2 Dimension convergence* In addition, we also extract features for each embedding dimension and include time-dependent information in them. The sequence reconstructed from this is as follows:

$$\mathcal{D}_t = (g_t^T W_g)^T \odot \text{sigmoid}(g_{t-1}^T W_g)^T \quad (8)$$

where $W_g \in \mathbb{R}^{3 \times 1}$, which extracts features from the embedding dimensions, sigmoid($\cdot$) is the activation function, which records historical information, and $\odot$ is the corresponding multiplication operation, which applies the recorded historical information to the current moment.

*3.3.3 decoder input* We concatenate the two feature sequences as $C = \text{Concat}(\mathcal{T}, \mathcal{D}) \in \mathbb{R}^{L_x \times (k+d_{model})}$ and multiply it by the fusion matrix $W_c \in \mathbb{R}^{(k+d_{model}) \times d_{model}}$ to get the regression input $\mathbf{X}_{reg} \in \mathbb{R}^{L_x \times d_{model}}$. This part of the sequences will be used in the decoder for multivariate regression. We feed the following sequences to the decoder:

$$\mathbf{X}_{rec} = \text{Concat}(\mathbf{X}_{reg}, \mathbf{X}_{pre}) \in \mathbb{R}^{(L_x + L_y) \times d_{model}} \quad (9)$$

where $\mathbf{X}_{pre} \in \mathbb{R}^{L_y \times d_{model}}$ is a placeholder for the target sequence (set scalar as 0).

## 4 EXPERIMENTS

### 4.1 Experimental setup

*4.1.1 Datasets* We conduct experiments on four large-scale datasets:
**Air Quality**[1]: This dataset contains hourly responses from a gas multi-sensor device deployed on site in an Italian city.
**Stock**[2]: This dataset tracks index daily price data from stock exchanges around the world (US, China, Canada, Germany, Japan, etc.)
**Electricity**[3]: This dataset contains the electricity consumption of 321 clients. Due to missing data, we converted the dataset to hourly consumption for 2 years.
**Smartphone**[4]: This dataset is an indoor positioning time series based on WLAN and geomagnetic field fingerprints.

*4.1.2 Baselines* We use classic models and state-of-the-art methods for evaluation: LSTM [4]: a variant of recurrent network with long and short-term memory. LST-Net [12]: a predictive model combining convolutional network and recurrent network. Reformer [7]: a transformer-based locality-sensitive hashing dot-product attention model. Informer [9]: an advanced model for long-term series forecasting. Autoformer [10]: an autocorrelation decomposition transformer for long-term sequence forecasting.

*4.1.3 Implementation details* We divide the dataset into training, validation, and test sets in a 6:2:2 ratio. To compare the performance of the model under different future horizons, we fix the input length to 96 and the prediction length to be {96, 192, 336, 720}. We use the Adam optimizer, the coefficient $\lambda$ of the regularization term is 0.01, the infinitesimal term $\varepsilon$ is 0.001, the batch size is 32, the number of epochs is 5, the initial learning rate is 5e-4, and the learning rate variation law is $lr_{epoch} = lr_0 \times \sum_{i=0}^{epoch} 0.9^i$. All experiments are performed on NVIDIA Tesla V100 32GB GPU.

---
[1] https://archive.ics.uci.edu/ml/datasets/Air+Quality
[2] https://www.kaggle.com/datasets/mattiuzc/stock-exchange-data
[3] https://archive.ics.uci.edu/ml/datasets/ElectricityLoadDiagrams20112014
[4] https://archive.ics.uci.edu/ml/datasets/GeoMagnetic+field+ans+WLAN+datasets+for+indoor+localisation+from+wristband+and+smartphone

**Table 1: Multivariate time series forecasting results on four datasets**

| Models | Metrics | Air Quality | | | | Stock | | | | Electricity | | | | Smartphone | | | | Count |
|---|---|---|---|---|---|---|---|---|---|---|---|---|---|---|---|---|---|---|
| | | 96 | 192 | 336 | 720 | 96 | 192 | 336 | 720 | 96 | 192 | 336 | 720 | 96 | 192 | 336 | 720 | |
| LSTM | MAE | 0.532 | 0.554 | 0.581 | 0.605 | 1.105 | 1.279 | 1.339 | 1.527 | 0.527 | 0.571 | 0.588 | 0.806 | 0.582 | 0.596 | 0.661 | 0.805 | 0 |
| | MSE | 0.496 | 0.533 | 0.540 | 0.632 | 1.220 | 1.447 | 1.866 | 2.984 | 0.475 | 0.515 | 0.574 | 0.780 | 0.640 | 0.673 | 0.844 | 0.969 | |
| LST-Net | MAE | 0.587 | 0.614 | 0.592 | 0.639 | 0.978 | 1.219 | 1.674 | 1.778 | 0.592 | 0.653 | 0.805 | 1.093 | 0.667 | 0.690 | 0.759 | 0.884 | 0 |
| | MSE | 0.596 | 0.639 | 0.608 | 0.764 | 1.115 | 1.671 | 2.117 | 3.211 | 0.534 | 0.592 | 0.912 | 1.874 | 0.805 | 0.834 | 0.967 | 1.225 | |
| Reformer | MAE | 0.643 | 0.625 | 0.681 | 0.693 | 1.413 | 1.447 | 1.464 | 1.442 | 1.013 | 1.169 | 1.255 | 1.281 | 0.930 | 1.198 | 1.243 | 1.479 | 0 |
| | MSE | 1.398 | 1.337 | 1.563 | 1.739 | 1.521 | 2.075 | 2.470 | 2.757 | 1.445 | 1.528 | 1.698 | 2.334 | 1.256 | 1.725 | 1.963 | 2.368 | |
| Informer | MAE | 0.443 | 0.396 | 0.437 | 0.477 | 0.578 | 0.654 | 0.694 | 0.812 | 0.435 | 0.449 | 0.515 | 0.503 | 0.369 | 0.415 | 0.422 | 0.391 | 0 |
| | MSE | 0.458 | 0.418 | 0.522 | 0.554 | 0.689 | 0.705 | 0.731 | 0.957 | 0.363 | 0.391 | 0.467 | 0.472 | 0.228 | 0.287 | 0.296 | 0.355 | |
| Autoformer | MAE | 0.377 | 0.398 | 0.402 | 0.393 | 0.350 | 0.432 | 0.540 | 0.478 | **0.386** | **0.432** | 0.497 | 0.517 | 0.338 | 0.396 | **0.370** | 0.495 | 4 |
| | MSE | 0.415 | 0.444 | 0.494 | 0.455 | 0.204 | 0.302 | 0.447 | 0.335 | 0.294 | 0.347 | 0.391 | 0.436 | 0.214 | 0.273 | **0.235** | 0.337 | |
| DRAformer | MAE | **0.270** | **0.325** | **0.339** | **0.332** | **0.137** | **0.171** | **0.239** | **0.260** | 0.460 | 0.487 | **0.447** | **0.463** | **0.306** | **0.351** | 0.392 | **0.387** | 28 |
| | MSE | **0.197** | **0.241** | **0.253** | **0.270** | **0.022** | **0.037** | **0.162** | **0.193** | **0.276** | **0.320** | **0.266** | **0.373** | **0.210** | **0.228** | 0.246 | **0.233** | |

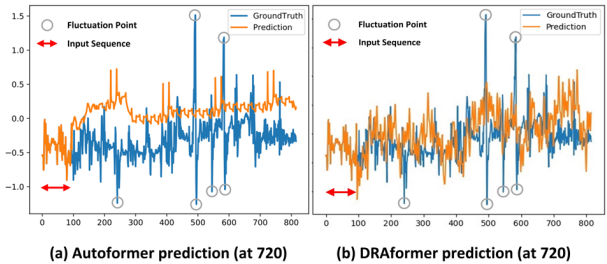

**Figure 2: Performance comparison of Autoformer and DRAformer on smartphone dataset**

## 4.2 Results Analysis

*4.2.1 comparative analysis* Table 1 shows the results of different prediction methods on the four datasets. We apply two evaluation metrics, MAE and MSE. As can be seen from Table 1, the comprehensive prediction ability of DRAformer is significantly higher than other baselines. Specifically, our model reduces MAE by an average of 54.1% (at 96), 53.0% (at 336) compared to LST-Net, and 29.4% (at 192), 25.9% (at 720) compared to Autoformer). In particular, our model has superior performance on the Stock dataset. It reduces MSE by 87.7% (at 192, compared to Autoformer), 94.8% (at 192, compared to Informer), and reduces MAE by 45.6% (at 720, compared to Autoformer), 68.0% (at 720, compared to Informer). Due to the obvious volatility of financial data, we believe that learning the difference series significantly improves the model's ability to tolerate disturbances. In addition, sensor data is also strongly sensitive to noise, as shown in Figure 2. We compare the performance of DRAformer and Autoformer on the smartphone dataset, and Figure 2 shows the predictions at 720 for both methods. The fluctuation point and input sequence are marked in the figure. It can be seen that our model has stable prediction performance and is not disturbed by outliers. Since the reconstructed attention captures the changing laws, it has a better predictive ability for data with strong variability. The autoformer is susceptible to outliers, its predicted trajectory deviates significantly from the true value, and it does not sufficiently learn the changing features of the sequence. In terms of long-sequence prediction, at a prediction length of 720, our model reduces MAE by an average of 56.2% and MSE by an average of 63.7%, indicating that our model can deeply capture long-term associations of sequences and has a stable prediction ability.

**Table 2: Ablation study on electricity dataset**

| Ablation item | 96 | | 192 | | 336 | | 720 | |
|---|---|---|---|---|---|---|---|---|
| | MAE | MSE | MAE | MSE | MAE | MSE | MAE | MSE |
| Rec Attn | 0.522 | 0.477 | 0.538 | 0.488 | 0.582 | 0.496 | 0.504 | 0.441 |
| Rec Seq | 0.465 | 0.355 | 0.552 | 0.402 | 0.545 | 0.459 | 0.502 | 0.438 |

*4.2.2 ablation analysis* We performed ablation analysis on DRAformer on the electricity dataset. We replace the reconstructed attention with multi-head attention and the reconstructed sequence with placeholders with location information, respectively. Table 2 shows the analysis results. It can be seen that the models lacking reconstructed attention have an average increase of 15.8% and 56.9% in MAE and MSE, and the models lacking reconstructed sequences have an average increase of 11.5% and 36.1% in MAE and MSE. This indicates that the capturing of changing features by attention in DRAformer contributes greatly to the prediction performance, and the reconstructed sequence further learns long-term dependencies on the basis of attention. When the prediction length is 720, the evaluation index decreases significantly once, indicating that our model has a stable ability to predict long sequences. In a word, the modules in our proposed model are indispensable.

## 5 CONCLUSION

In this paper, we propose DRAformer for time series forecasting. Specifically, our proposed reconstructed attention method learns hidden changing features by paying attention to the distance and probability differences of the sequences, and the reconstructed feature sequences preserve more time-dependent information. Experiments on real datasets demonstrate the superior performance of our model. In the future, we will further investigate the efficient learning of sequence variation associations in the context of high-dimensional data.

# REFERENCES

[1] Mehrmolaei, Soheila, and Mohammad Reza Keyvanpour. Time series forecasting using improved ARIMA. In 2016 Artificial Intelligence and Robotics (IRANOPEN), pp. 92-97. IEEE, (2016).

[2] Sapankevych, Nicholas I., and Ravi Sankar. Time series prediction using support vector machines: a survey. IEEE computational intelligence magazine 4, no. 2 (2009): 24-38.

[3] Hill, Tim, Marcus O'Connor, and William Remus. Neural network models for time series forecasts. Management Science 42, no. 7 (1996): 1082-1092.

[4] Hochreiter, Sepp, and Jürgen Schmidhuber. Long short-term memory. Neural computation 9, no. 8 (1997): 1735-1780.

[5] Cho, Kyunghyun, Bart Van Merriënboer, Caglar Gulcehre, Dzmitry Bahdanau, Fethi Bougares, Holger Schwenk, and Yoshua Bengio. Learning phrase representations using RNN encoder-decoder for statistical machine translation. arXiv preprint arXiv:1406.1078 (2014).

[6] Vaswani, Ashish, Noam Shazeer, Niki Parmar, Jakob Uszkoreit, Llion Jones, Aidan N. Gomez, Łukasz Kaiser, and Illia Polosukhin. Attention is all you need. Advances in neural information processing systems 30 (2017).

[7] Kitaev, Nikita, Łukasz Kaiser, and Anselm Levskaya. Reformer: The efficient transformer. arXiv preprint arXiv:2001.04451 (2020).

[8] Li, Shiyang, et al. Enhancing the locality and breaking the memory bottleneck of transformer on time series forecasting. Advances in Neural Information Processing Systems 32 (2019).

[9] Zhou, Haoyi, et al. Informer: Beyond efficient transformer for long sequence time-series forecasting. Proceedings of AAAI.(2021).

[10] Xu, Jiehui, Jianmin Wang, and Mingsheng Long. Autoformer: Decomposition transformers with auto-correlation for long-term series forecasting. Advances in Neural Information Processing Systems 34 (2021).

[11] Zhou, Tian, Ziqing Ma, Qingsong Wen, Xue Wang, Liang Sun, and Rong Jin. FEDformer: Frequency enhanced decomposed transformer for long-term series forecasting. arXiv preprint arXiv:2201.12740 (2022).

[12] Guokun Lai, Wei-Cheng Chang, Yiming Yang, and Hanxiao Liu. Modeling long- and short-term temporal patterns with deep neural networks. In SIGIR, (2018)